# Bangla Text Recognition from Video Sequence: A New Focus

Souvik Bhowmick[#1], Purnendu Banerjee[#2]

*Computer Vision and Pattern Recognition Unit, Indian Statistical Institute,*

*203 B. T. Road, Calcutta 700 035, India*

[1]sou.bhowmick@gmail.com
[2]purnendubannerjee@yahoo.com

*Abstract*— extraction and recognition of Bangla text from video frame images is challenging due to complex color background, low-resolution etc. In this paper, we propose an algorithm for extraction and recognition of Bangla text form such video frames with complex background. Here, a two-step approach has been proposed. First, the text line is segmented into words using information based on line contours. First order gradient value of the text blocks are used to find the word gap. Next, a local binarization technique is applied on each word and text line is reconstructed using those words. Secondly, this binarized text block is sent to OCR for recognition purpose.

## 1. INTRODUCTION

Though lots of content based retrieval algorithms are developed for meeting requirements of text indexing and retrieval from video in the field of image processing and multimedia, recognition of Bangla text from low contrast video images is still an open challenge. With the proliferation of videos on the air, there is an increasing demand for search and retrieval of Bangla textual information. Besides, the current Bangla OCR has limitation that it segments words and characters well for high contrast and plane background document images but not for video images. This makes segmentation and recognition of Bangla video text more challenging and interesting.

We are concerned here with the recognition of *Bangla*, the second most popular script and language in the Indian sub-continent. About 200 million people of Eastern India and Bangladesh use this language, making it the fourth most popular in the world. A few reports are available on the recognition of isolate Indian script character ([4]-[6]) but none deal with a complete OCR for complex background Bangla text video images. From that standpoint this paper is a pioneering work among all Indian scripts after Tiwari [7] et. al. This paper will be useful for the computer recognition of several other Indian script forms since these scripts having the same ancient origin.

Monitoring of videos with textual information has been an important task for media analysts, business world, intelligence agencies etc. The reports generated by monitoring agencies can be used to have diverse perception on similar incidents. One way is to manually monitor the video round the clock and then produce it for analysis. However manual monitoring is not only requires a lot of concentration but also an expertise in the languages in which the text in videos is being broadcasted. It is also possible that the reports generated by human monitors can be biased, based on their own perception and may vary from person to person. So informative videos monitored manually from multiple sources may not be a good option. To overcome this, automation of video text analysis is required.

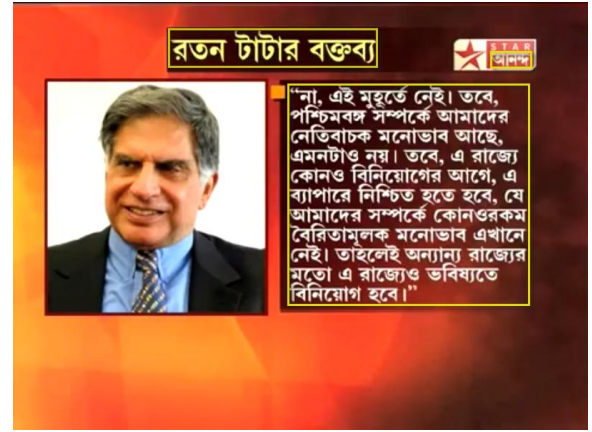

Fig. 1 Bangla text detected on a video sample image

According to [1], a video text detection and recognition system consists of five steps: (1) detection, (2) localization, (3) tracking, (4) extraction, and (5) Recognition. The first three steps focus on detecting and localizing text lines in video frames. The text lines' locations are usually represented by their rectangular bounding boxes. However, these boxes still contain both text and background pixels so the fourth step aims to make text easier to Recognize, e.g. by binarization. The fifth and final step typically uses an optical character recognition (OCR) engine to produce the final output as a string of characters. OCR engines work well for document images, most of which contain monochrome text on a plain background. However, it does not produce satisfactory results for video Images due to the poor resolution, low contrast and unconstrained background of the text lines. Moreover, scene Text is affected by lighting conditions and perspective Distortions. Therefore, the character quality in a video frame is too low to be processed by a conventional OCR system directly. In addition, popular lossy compression methods such as MPEG often lower the image quality even further. Hence, in this paper, we use our text detection Method [2] to get text lines in a video frame and focus on the fourth step, extraction and the fifth one, recognition.

A common video character segmentation method is projection profile analysis [3]. Edge information (or other kinds of energy) in each column was analyzed to distinguish between columns that contain text pixels and gap columns, based on the observation that the former had higher energy compared to the latter. Heuristic rules were proposed to further split and merge the segmented regions based on assumptions about the characters' widths and heights.

## 2. PROPOSED METHODOLOGY

Since our focus is on Bangla word and character extraction and recognition from video text lines that are detected by the text detection method proposed in [2] for video text location with clear bounding box. The reason for choosing this method is that the method is able to locate both graphics and scene text and multi-oriented text in complex video background despite its low resolution.

The proposed method is structured into six subsections. The Absolute Gradient features are explained in Subsection 2.1 with illustration for extracting text pixels using K-Means clustering concept. In Subsection 2.2, we propose a method for word extraction using the Vertical Projection concept. Subsection 2.3 describes the methodology to detect the Matra and separation of three different zones of Bangla text line. Based on this Zone information of the text line, the proposed approach to segment the words into characters are detailed in subsection 2.4. In subsection 2.5, the feature vectors to train the SVM (Support Vector Machine) is mentioned and explained. The Recognition procedure of Basic and Compound Bangla Characters are described in subsection 2.6.

### 2.1 Absolute Gradient Feature

Since video images have low resolution and complex background, we need some mechanism to differentiate low contrast text pixels from background. So we are proposing the **Absolute Gradient Feature** method, applying which we can get the edge information for the textual area over the image. For a given text line image, we are using the **Horizontal Gradient Information** to get high frequency components for text pixels. We are using the **[-1 1] mask**ing here to get the Gradient information corresponding to each pixel in the original grayscale image. In the terms of mathematics,

$$grad_x(x, y) = |img(x+1, y) - img(x, y)|$$

Again **Vertical Gradient Information** are calculated by using **(-1 1) mask** to each pixel of the original grayscale image.

$$grad_y(x, y) = |img(x, y+1) - img(x, y)|$$

TABLE 1
Resultant Image after Absolute Gradient Feature

| Original Grayscale Image | Normalized Gradient Image |
|---|---|
| 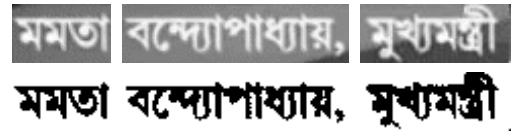 | |

After getting this absolute gradient image we use **K-Means clustering** to binarize the image.

TABLE 2
Resultant Image after K-Means Clustering

| Normalized Gradient Image | K-Mean Cluster(Binarized) Image |
|---|---|

### 2.2 Word Segmentation

K-Mean clustering with two classes, gives a text cluster as a binarized edge map for the whole text line. A component analysis is then driven out on the Text cluster to get the average area for the components. Now those components which have area less than two percent of the average area are ignored as noise. The pink-colored components in fig-2 (A) are marked as the noise and then removed as fig-2 (B)

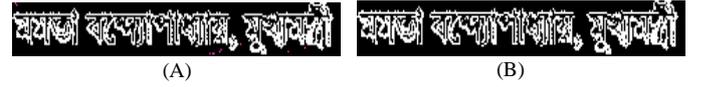

(A)     (B)
Fig. 2 Noise identification on Text Cluster

Then for the resultant text cluster, we are projecting it vertically to get some continuous zero-frequency columns. Those continuous intervals are marked as word gaps. The Gaps detected on a text cluster can be seen in the fig-3

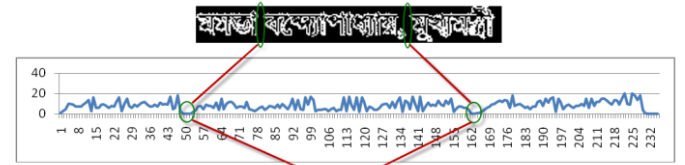

Fig. 3 Vertical Projection to detect word gaps

Based on the Word Gap information, the original grayscale Text line image is now divided into some word images. Each word grayscale image is now binarized using **Otsu's Thresholding** technique [3].

Fig. 4 Binarization of word images

Now merging these binarized word images, we are getting the whole binarized text line back. This binarized text line will be moved as the input to the next step of our algorithm.

### 2.3 Matra detection and Zone Separation

Since *Bangla* text lines can be partitioned into three zones (see Fig. 5), it is convenient to distinguish these zones. Character recognition becomes easier if the zones are distinguished because the lower zone contains only modifiers, while the upper zone contains modifiers and portions of some basic characters.

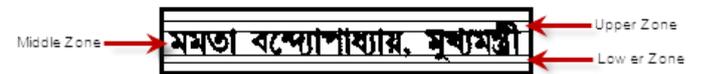

Fig. 5 Different zones of Bangla Text

Firstly a horizontal projection is applied on the binarized image to find the span of its highest peak, which is referred as the *Matra* of the text line (shown in fig-6).

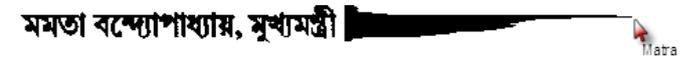

Fig. 6 Matra Detection

From Fig. 5 it is clear that the upper zone can be separated from the middle zone of a text line by the Matra, which is already detected. To separate the middle and lower zone consider an imaginary straight line *xy* that horizontally partitions the text line region into equal halves. Consider only the connected components below *xy* and connected to *xy*. The horizontal line, which passes through maximum number of the lowermost pixels of each of these components, is the separator line (**Base line**) between the middle and lower zones. See Fig- 7 for an illustration

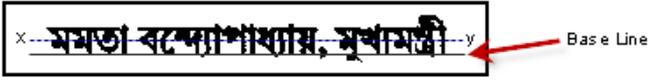

Fig. 7 Lower boundary detection

## 2.4 Character Segmentation

To segment individual characters of a word we consider only the middle zone. The basic approach here is to ignore the *Matra* so that the characters get topologically disconnected, as shown in Fig. x. To find the demarcation line of the characters a linear scanning in the vertical direction from the Base line is initiated. If during a scan, one can reach the Matra without touching any black pixel then this scan marks a boundary between two characters. For some *kerned* characters (24) a piecewise linear scanning method has been invoked (see Fig. 8). Using this approach nearly 92.6% characters have been properly segmented. The error was due to touching characters arising out of the binarization of the video images.

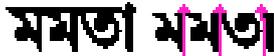

Fig. 8 Character segmentation after deletion of Matra

## 2.5 Primary Grouping Of Characters

Before going for actual classification, the basic, modified and compound characters are distinguished. One motivation of this step is that although there are a large number of compound characters, they occupy only 4% to 6% of the text corpus. If we can isolate them, then the small sub-set of basic characters can be recognized by a fast, accurate and robust technique while we could use a slower technique to recognize the compound characters. We have used a feature-based approach for basic and modifier character recognition, and a combination of feature-based and template- matching approach for the compound character recognition. The primary grouping of characters is done with the following observations. The middle zone of a script line may contain the vowel modifiers that have very small width. The width of the basic characters vary from character to character, being the least for character D (দ) and the most for character N+ (ঞ) in almost all fonts. On the other hand, some compound characters like /DG/ (দ্গ), /R,G/ (ক্গা) have very large width. Another important point to note is that the compound characters are, in general, more complex shaped than the modifier and basic characters. In general, a compound character will have large border length and high curvature along the border than a basic character. Thus, using the following three features a character may be identified as belonging to one of the three groups representing modifier, basic and compound characters.

- Feature *f*1. Width-height ratio of middle zone
- Feature *f*2. Normalized height of the leftmost black pixel
- Feature *f*3. Normalized distance of the lowermost black pixel from left point of bounding box
- Feature *f*4. Normalized distance of the rightmost black pixel from top
- Feature *f*5. Length of *Matra*
- Feature *f*6. No. of touching lines to *Matra* region
- Feature *f*7. Length of longest vertical line & its distance from left of bounding box (both normalized).
- Feature *f*8. The Contour image of every Character patch is divided into some 5X5 subparts. A 4-directional feature is then calculated on each of those sub-parts, which means we are calculating a feature vector consisting of 100 (=5X5X4) dimension.

Two discriminant planes of the form

$a1f1+a2f2+a3f3+a4f4+a5f5+a6f6+a7f7+a8f8=a0$ and
$b1f1+b2f2+b3f3+b4f4+b5f5+b6f6+b7f7+b8f8=b0$

are used for the classification. The parameters of discriminant planes are found from a large number of training data. For 12 point text digitized at 300 dpi the typical values of $a0, a1, a2, a3,…$ and $b0, b1, b2, b3,…$ are 225.77, 0.30, 0.21, 1.99,… and 80.83, 0.22, 0.49, 0.67,… respectively. On the test data set, an overall classification accuracy of 81.5% was obtained by this classifier.

## 2.6 Recognition of Characters

### A) Basic character and modifier recognition

Our recognition scheme of modifier and basic characters is based on shape identification only. A feature based tree classifier separates the characters. For the compound characters, feature-based tree classifier is initially used to separate them into small groups. Next, an efficient template matching approach is employed to recognize individual characters of each group.

### B) Compound character recognition

The compound character recognition is done in two stages. In the first stage the characters are grouped into small sub-sets by a feature-based tree classifier. At the second stage characters in each group are recognized by a sophisticated run-based template matching approach. We adopted this hybrid approach instead of only tree classifier approach because it is nearly impossible to find a set of stroke features which are simple to compute, robust and reliable to detect and are sufficient to classify a large number of complex-shaped compound characters. The features used in the tree classifier are head line, vertical line, left slant, boundary box width, presence of signature in upper zone etc. A terminal node of this tree corresponds to a sub-set of about 20 characters. These character templates are ranked in terms of their bounding box width and stored during the training phase of the classifier at an address corresponding to this terminal node. A candidate compound character which has reached a terminal node of the tree is then matched against the stored templates corresponding to this node. Matching is started with the template whose bounding box width is nearest to that of the candidate and continued till the template width is comparable to that of the candidate. To superimpose the candidate on the template for computing the matching score, the reference position is chosen by characteristic features of the candidate.

## 3. EXPERIMENTAL RESULT

For the experimentation purpose we have created a dataset consisting of 650 text lines, and 1300 words and 4120 characters. To evaluate performance of the proposed method, we mainly consider the recognition rate of Bangla characters based on the true positive and false positive cases on our given dataset.

The sample results for word and character extraction given by the proposed method is shown in Tables 3a, 3b and 4 where gaps are shown in white color for different varieties of word and character images.

TABLE 3a
Sample results for word extraction

| Original Grayscale | Gradient Edge Image |
|---|---|
| স্বায়ত্বশাসনের ভাবনা | স্বায়ত্বশাসনের ভাবনা |
| মুকুল রায়, জাহাজ প্রতিমন্ত্রী | মুকুল রায়, জাহাজ প্রতিমন্ত্রী |
| পুলিশ পোশাকে 'ডাকাত' | পুলিশ পোশাকে ডাকাত |
| নিরাপত্তায় বিশেষ নজর | নিরাপত্তায় বিশেষ নজর |
| জয়ন্ত হাটি, স্থানীয় বাসিন্দা | জয়ন্ত হাটি, স্থানীয় বাসিন্দা |
| দুর্ঘটনার বলি ৬ | দুর্ঘটনার বলি ৬ |
| গ্রামি পুরস্কার জোবসকে | গ্রামি পুরস্কার জোবসকে |
| পুলিশের অনুমতি | পুলিশের অনুমতি |
| নিরাপত্তায় নজর | নিরাপত্তায় নজর |
| পুলিশ সেজে ডাকাতি | পুলিশ সেজে ডাকাতি |

TABLE 3b
Binarized words for each Text line

| Binarized Text Lines |
|---|
| স্বায়ত্বশাসনের ভাবনা |
| মুকুল রায়, জাহাজ প্রতিমন্ত্রী |
| পুলিশ পোশাকে 'ডাকাত' |
| নিরাপত্তায় বিশেষ নজর |
| জয়ন্ত হাটি, স্থানীয় বাসিন্দা |
| দুর্ঘটনার বলি ৬ |
| গ্রামি পুরস্কার জোবসকে |
| পুলিশের অনুমতি |
| নিরাপত্তায় নজর |
| পুলিশ সেজে ডাকাতি |

TABLE 4
Sample results for character extraction

| Binarized Word | Character Gap detected | Recognized Characters |
|---|---|---|
| গ্রামি | গ্র া মি | গ্রামি |
| পুরস্কার | প ু র স্ক া র | পুরস্কার |
| পুলিশের | প ু ল ি শ ে র | পুলিশেব |
| অনুমতি | অ ন ু ম তি | অনুমতি |
| নিরাপত্তায় | নি র া প ত্তা য় | নিরপতায় |
| বিশেষ | বি শে ষ | বিশেষ |
| জয়ন্ত | জ য ন্ত | জয়ন্ত |
| দুর্ঘটনার | দ ু র্ঘ ট না র | দুঘটনার |
| সেজে | সে জে | সেজে |
| মমতা | ম ম তা | মমতা |
| মুখ্যমন্ত্রী | ম ু খ্য ম ন্ত্রী | মুখামন্ত্রী |

TABLE 5
Recognition Result

| Total No. of Words | Total No. of Characters | Character Recognized | Recognition Rate |
|---|---|---|---|
| 1300 | 4120 | 3360 | **81.5%** |

## 5. CONCLUSION

In this work, we have proposed a new and novel method of segmentation based on Gradient features for word and character extraction from Bangla text line and word patch images.

The main future scope of our work is to rise up the recognition rate of the system higher by introducing some more robust feature for the Bangla Compound characters. Since in video images the resolution is so low, recognizing compound characters becomes more and more difficult. So, that will be the main point of interest for further modification in near future.